%% file: icphs2023-template.tex
\documentclass[a4paper,11pt,twocolumn]{article}

\usepackage{icphs2023}
\usepackage{epstopdf}

\usepackage{threeparttable}
\usepackage{amsmath} %asmath
\usepackage{cleveref} % cref
\usepackage{breqn}
\usepackage{amssymb} %square
\usepackage[extra]{tipa}
\usepackage{subfig}
\usepackage{url}
\usepackage{multirow}
\usepackage{kotex}

\title{Comparison of L2 Korean pronunciation error patterns from five L1 backgrounds \\ by using automatic phonetic transcription}

\author{Eun Jung Yeo\footnote{Equal contributions.}, Hyungshin Ryu*, Jooyoung Lee, Sunhee Kim, Minhwa Chung}
\organization{Seoul National University}
\email{\{ej.yeo, rhss10, excalibur12, sunhkim, mchung\}@snu.ac.kr}
\begin{document}

\maketitle
\input{sections/0_abstract}
\input{sections/1_introduction}
\input{sections/2_method}
\input{sections/3_result}
\input{sections/4_discussion}
\input{sections/5_conclusion}
\section*{Acknowledgment}
\vspace{-2mm}
This research was supported by the MSIT (Ministry of Science and ICT), Korea, under the ITRC (Information Technology Research Center) support program (IITP-2023-2018-0-01833) supervised by the IITP (Institute for Information\&Communications Technology Planning\&Evaluation)

\bibliographystyle{IEEEtran}
\bibliography{icphs2023}

\theendnotes

\end{document}

%% file: sections/0_abstract.tex
\begin{abstract}
This paper presents a large-scale analysis of L2 Korean pronunciation error patterns from five different language backgrounds, Chinese, Vietnamese, Japanese, Thai, and English, by using automatic phonetic transcription. For the analysis, confusion matrices are generated for each L1, by aligning canonical phone sequences and automatically transcribed phone sequences obtained from fine-tuned Wav2Vec2 XLS-R phone recognizer. Each value in the confusion matrices is compared to capture frequent common error patterns and to specify patterns unique to a certain language background. Using the Foreign Speakers' Voice Data of Korean for Artificial Intelligence Learning dataset, common error pattern types are found to be (1) substitutions of aspirated or tense consonants with plain consonants, (2) deletions of syllable-final consonants, and (3) substitutions of diphthongs with monophthongs. On the other hand, thirty-nine patterns including (1) syllable-final \textipa{/l/} substitutions with \textipa{/n/} for Vietnamese and (2) \textipa{/\textturnm/} insertions for Japanese are discovered as language-dependent. 
\end{abstract}

\begin{keywords}
Comparative analysis, Pronunciation error patterns,  Automatic phonetic transcription, L1 influence, L2 Korean speech
\end{keywords}

%% file: sections/1_introduction.tex
\section{Introduction} \label{sec:intro}
\input{figures/diagram.tex}

Computer Assisted Pronunciation Training (CAPT) has emerged as an effective tool for non-native speakers, offering cost-effective feedback while overcoming the time and location constraints of traditional language learning \cite{rogerson2021computer, chao20223m}. 
With the advancement of deep learning, novel architectures have been proposed to improve the detection and diagnosis performance of the CAPT system \cite{chao20223m,w2v_momentum,peng_mdd}.
Linguistic studies were conducted with a different point of view, where the characteristics of non-native speech are explored \cite{eng_l2, egypt_l2, french_align, eng_align, nasal_dnn, pca_icphs, pca_interspeech}. 

The influence of the first language (L1) on the second language (L2) pronunciation has long been investigated.
However, due to the requirement of time- and labor-intensive human transcription, analyses were often limited to certain acoustic phenomena from a small number of participants.
For example, 127 obstruent-initial words were collected from 22 English-speaking learners to analyze their realizations of Korean stops \cite{eng_l2}, while 11 words with monophthongs or diphthongs were recorded from 53 Egyptians to understand their Korean vowel perception and production \cite{egypt_l2}.
On the other hand, some studies have taken advantage of deep learning methods including forced alignments \cite{french_align, eng_align} or feature extraction \cite{nasal_dnn, pca_icphs, pca_interspeech}, which allowed an analysis of larger materials or participants.
Nevertheless, their analyses were often limited to a small subset of phone inventories, such as fricatives \cite{french_align}, nasals \cite{nasal_dnn}, and vowels \cite{eng_align, nasal_dnn, pca_icphs, pca_interspeech}. 

To mitigate the limitation of human transcriptions especially on large-scale analysis, utilizing automatic transcription can be one possible option.
Model-based transcriptions have advantages in terms of time- and cost-efficiency, consistency, and objectivity \cite{cucchiarini2003automatic}.
With self-supervised-based speech models (SSL) showing state-of-the-art performances in various speech tasks including automatic speech recognition \cite{babu2021xls, multilingual_asr}, this work proposes to employ automatic phonetic transcription for L2 mispronunciation analysis.

This study conducts a large-scale analysis on L2 Korean pronunciation error patterns of speakers from five different L1 backgrounds, Chinese, Vietnamese, Japanese, Thai, and English.
Confusion matrices are generated by aligning the canonical and automatically transcribed phones obtained from fine-tuned SSL Wav2Vec2 XLS-R recognizer \cite{babu2021xls}. 
The error patterns for each non-native speech are inferred from the confusion matrix, and are scrutinized with respect to L1 background.
We expect our analysis to bring new insights into L1-influenced pronunciations and provide linguistic clarity in developing the CAPT system.

%% file: figures/diagram.tex
\begin{figure*}[h!] % t: top, b: bottom, h: here
\centering
\includegraphics[width=1\textwidth]{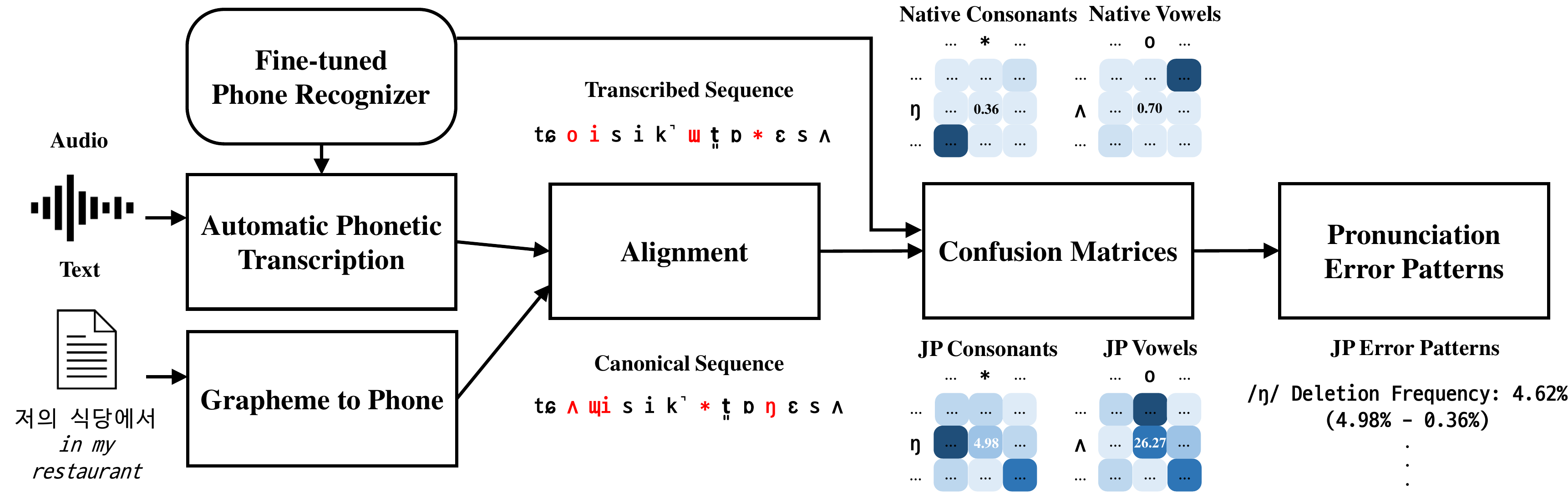}
\caption{Overview of the pronunciation error patterns analysis using automatic phonetic transcription}

\label{fig:diagram}
\end{figure*}

%% file: sections/2_method.tex
\section{Method} \label{sec:method}
\subsection{Corpus} \label{ssec:Corpus}
Foreign Speakers' Voice Data of Korean for Artificial Intelligence Learning \cite{l2koreandataset} is an open-source dataset designed to train automatic speech recognition systems for Korean learners. 
The dataset consists of L2 Korean speech from 1911 speakers with 80 different L1s, where each speaker recorded both read-speech and spontaneous speech.
In this paper, we restricted the analysis to five L1 backgrounds, where the most abundant recordings were collected - Chinese (ZH), Vietnamese (VI), Japanese (JP), Thai (TA), and English (EN).
In addition, we examined only the read-speech to exclude the influence of speech fluency and focus on pronunciation error patterns. Participants without official scores on the Test of Proficiency in Korean (TOPIK) are also excluded from the analysis. As \Cref{tbl:dataset} presents, the dataset includes beginners, intermediate and advanced learners for all five language backgrounds.

\input{figures/datasets.tex}
\subsection{Analysis} \label{ssec:calculation}
\Cref{fig:diagram} is the overview of our analysis. First, \textbf{(1) automatic phonetic transcription} on non-native speech is performed by using the fine-tuned Wav2Vec2 XLS-R phone recognizer. Next, \textbf{(2) confusion matrices} for both native and non-native speech are created by aligning the canonical phone sequences and automatically transcribed phone sequences.
Finally, \textbf{(3) pronunciation error patterns} for non-native speech are extracted from confusion matrices and analyzed regarding five L1 backgrounds. We release the source code and the fine-tuned model for the ease of reproduction.\footnote{https://github.com/eunjung31/L2K-PronunciationError}

\vspace{-2mm}
\subsubsection{Automatic phonetic transcription} \label{ssec:recognizer}
For automatic phonetic transcription, 
we choose to utilize Wav2Vec2 XLS-R-300m model, which is pre-trained on 128 different languages including Chinese, Vietnamese, Japanese, Thai, and English \cite{babu2021xls}.
The model is fine-tuned with a private, phonetically balanced native Korean read-speech corpus, with a split of 108 hours of the training set (54000 audio files) and 12 hours of the test set (6000 audio files). 
The transcription used for fine-tuning is generated by trained linguists with the phone set introduced in
 \cite{mcauliffe2017montreal}. Thus the same set is used for error pattern analysis in this study. The phone set includes 40 phones, consisting of 19 consonants, 17 vowels, and 4 allophones. As for allophones, \textipa{[p\textcorner]},\textipa{[t\textcorner]},\textipa{[k\textcorner]} are stops at codas and \textipa{[\textfishhookr]} is \textipa{/l/} in syllable-initial position.
Our model showed a Phone Error Rate (PER) of 3.88\% on the test set, which verifies the model's performance. 

\vspace{-2mm}
\input{figures/confusionMatrix.tex}
\subsubsection{Confusion matrices} \label{ssec:confusion}
Confusion matrices are created for the native dataset, the non-native dataset as a whole, and each L1 group. Canonical phone sequences are generated with the grapheme to phone module in Montreal Forced Aligner \cite{mcauliffe2017montreal} and are aligned with the decoded phone sequences from the fine-tuned model, by using SCTK package offered at Kaldi \cite{povey2011kaldi}. \Cref{fig:jp} demonstrates the vowel confusion matrix for Japanese L2 Korean speech which refers to L2 error patterns. * represents insertion and deletion. Considering each value as the percentage of observations of the canonical phone row, each row sums to 100.

\subsubsection{Pronunciation error patterns} \label{ssec:errorMethod}
Pronunciation error patterns for each L1 background are inferred from the corresponding confusion matrix. Error patterns include substitution, insertion, and deletion errors. We propose two types of analyses to scrutinize L1-influenced error patterns.

In the first analysis, we focus on determining the common error patterns shared across five non-native groups. These common patterns can reveal the general characteristics of L2 Korean speech and be used for core feedback in pronunciation training. The first step involves choosing phones for each L1 background that have an accuracy lower than the average accuracy. Different thresholds are used because the performance of the phone recognizer can vary depending on the L1 background. Then, a list of pronunciation error patterns is created for each L1 background, which includes the three most common error patterns for each selected phone. Lastly, the common error patterns are identified by comparing the five error pattern lists and determining which error patterns are shared.

As for the second analysis, we aim to find unique patterns for each L1. Language-dependent error patterns can give guidelines for providing learners with customized feedback. The Chi-square test of independence is conducted to compare each L1 group to the average of five L1 groups to look at the association of L1 with each error pattern. 
The four most frequently observed patterns for each canonical phone are only tested, leading to a total of 840 error patterns (5 L1 backgrounds $\times$ 42 canonical phones (40 phones, insertion, deletion) $\times$ 4 realized phones) being statistically analyzed.

For both analyses, native confusion matrix error values were subtracted from corresponding non-native confusion matrix values beforehand, as they include not only the errors of the phone recognizer itself but also the pronunciation variation of native speakers.

%% file: figures/datasets.tex
\vspace{3mm}
\begin{table}[h!]
\caption{
Number of speakers and utterances
}
\label{tbl:dataset}

% \scriptsize
\centering
\resizebox{0.45\textwidth}{!}
{

\begin{tabular}{c|c|c|c|c|c|c|c}
\hline
\multirow{2}{*}{Language} & \multicolumn{2}{c|}{Beginner} & \multicolumn{2}{c|}{Intermediate} & \multicolumn{2}{c|}{Advanced} & \multirow{2}{*}{Total Hrs} \\
\cline{2-7}
& spk & utt & spk & utt & spk & utt &  \\
\hline
Chinese  & 2 & 513 & 56 & 18599 & 291 & 127811 & 487.41 h \\
\hline
Vietnamese  & 12 & 2837 & 85 & 36100 & 144  & 74921 & 412.91 h  \\
\hline
Japanese  & 3 & 1474 & 27 & 12004 & 140 & 87915 & 326.35 h \\
\hline
Thai  & 44 & 19773 & 57 & 26377 & 44 & 26688 & 260.57 h  \\
\hline
English & 14 & 938 & 15 & 1682  & 12 & 4272 & 24.34 h \\
\hline

\end{tabular}

}
\end{table}

%% file: figures/confusionMatrix.tex
\begin{figure}[t!] % t: top, b: bottom, h: here
\centering
\includegraphics[width=0.47\textwidth]{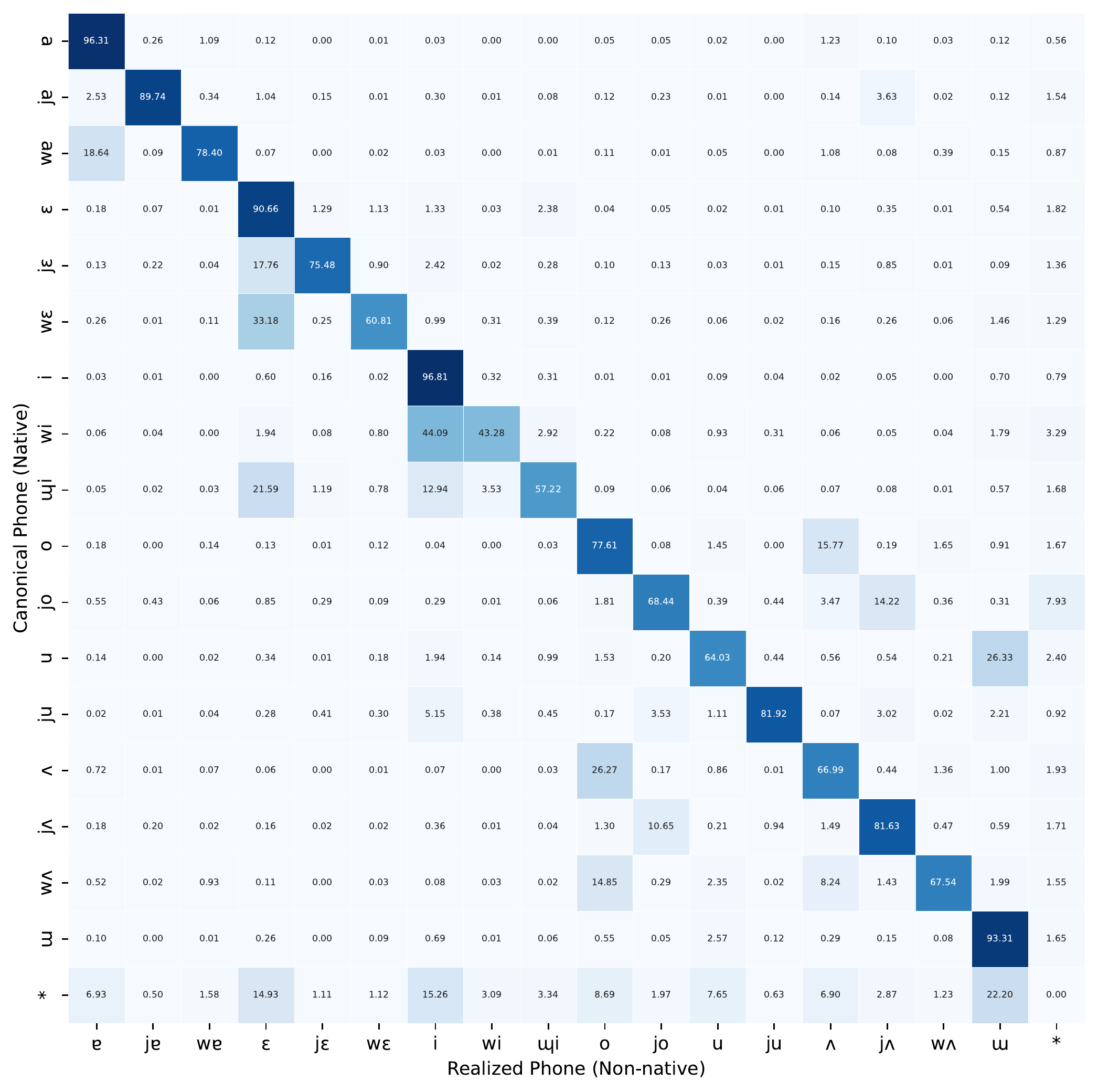}
\caption{Confusion matrix for L2 Korean vowel pronunciation of Japanese speakers}
\vspace{3mm}
\label{fig:jp}
\end{figure}

%% file: sections/3_result.tex
\section{Results} \label{sec:result}

\subsection{Common error patterns} \label{ssec:L1common}
Three types of error patterns are observed to be commonly shared across five languages: two types for consonants and one type for vowels. 
For consonants, seven error patterns are related to the first type, substitution errors of tense or aspirated with plain consonants: \textipa{/t\textsuperscript{h}/}to\textipa{/t/},
\textipa{/k\textsuperscript{h}/}to/k/,
\textipa{/\texttctclig\textsuperscript{h}/}to\textipa{/\texttctclig/};
\textipa{/\subdoublevert{p}/}to\textipa{/p/},
\textipa{/\subdoublevert{t}/}to\textipa{/t/},
\textipa{/\subdoublevert{\texttctclig}/}to\textipa{/\texttctclig/},
\textipa{/\subdoublevert{s}/}to\textipa{/s/}. 
The second error type is deletion errors of syllable-final, which includes \textipa{[t\textcorner]} and \textipa{/l/} coda deletions. 
As for vowels, substitution errors from diphthongs to monophthongs are found: \textipa{/wi/}to\textipa{/i/}, \textipa{/\textturnmrleg i/}to\textipa{/i/}, \textipa{/w\textturnv/}to\textipa{/\textturnv/}. 

\subsection{Language-dependent error patterns} \label{ssec:L1depend}
According to the statistical results, only 39 out of 840 pronunciation patterns showed L1 association.
25 patterns had L1 causing speakers with higher mispronunciation rates than the average, and the rest 14 had L1 being helpful to better pronunciations.
In this paper, we focus on the former 25 pronunciation error patterns as presented in \Cref{tab:compare} and show the full results in our repository$^{1}$. \textit{Canon} and \textit{Real} each refers to canonical and realized phones, with \textit{Average Freq} referring to the confusion matrix value averaged across the five L1s. For example, row 1 can be interpreted that the \textipa{[k\textcorner]} accuracy of Chinese learners was lower than the average of five L1s.

\textbf{Chinese}-speaking learners have 3 language-dependent error patterns including \textipa{[k\textcorner]} mispronunciation, \textipa{[k\textcorner]} deletion, and \textipa{/ju/}to\textipa{/jo/} substitution. \textbf{Vietnamese}-speaking learners have 3 error patterns which includes substitution from \textipa{/\subdoublevert{\texttctclig}/}to\textipa{/\texttctclig\textsuperscript{h}/}, \textipa{/l/} mispronunciation, and the following \textipa{/l/}to\textipa{/n/} substitution. \textbf{Japanese}-speaking learners  have the most various association with L1, with 18 errors showing statistical significance. The error patterns include mispronounced aspirated stops (\textipa{/p\textsuperscript{h}/},\textipa{/t\textsuperscript{h}/},\textipa{/k\textsuperscript{h}/}) and their substitution to tense stops 
(\textipa{/\subdoublevert{p}/},\textipa{/\subdoublevert{t}/},\textipa{/\subdoublevert{k}/}).
Mispronunciation of \textipa{/\ng/} and its substitution to \textipa{/n/} are more frequent than average learners. Insertion of \textipa{/\textfishhookr/} and \textipa{/\textturnm/} and substitution of diphthongs to monophthongs are also observed to be significantly frequent (\textipa{/w\textturna/,/w\textepsilon/,/wi/}). Lastly, the learners often mispronounce monophthongs (/\textipa{u}/,/\textipa{\textturnv}/) to different monophthongs (/\textipa{\textturnm}/,/\textipa{o}/). For \textbf{Thai}-speaking learners, frequent substitution of \textipa{/\texttctclig\textsuperscript{h}/}to\textipa{/s/} are found. No phones are significantly mispronounced by \textbf{English}-speaking learners only. 
This result may imply that English-speaking learners have advantages in learning Korean compared to other learners.
However, to validate this claim, further research using a more balanced dataset is essential, considering the limited number of English-speaking participants in our dataset.

%% file: sections/4_discussion.tex
\section{Discussion} \label{sec:Discussion}
Based on the observations in \Cref{sec:result}, we uncover the three L1-influence factors attributing to the L2 Korean pronunciation error patterns. The result of our analysis is found to be further supported by the conclusions from various previous literature. 

First is differences in \textbf{phoneme inventory} \cite{wang2008mispronunciation, khanal2021articulatory}. Substitution errors of three-way contrasted consonants can be explained by languages not distinguishing consonants as plain-tense-aspirated \cite{holliday2015longitudinal}. The commonly found substitutions to plain consonants further support the analysis, regarding each five languages includes plain consonants. Results also reveal an interesting finding of Japanese learners' consistent tensification of aspirated stops, alongside the tendency to substitute aspirated stops to plain. Previously, word-medial tense and aspirate distinction in perception have been reported to be difficult for Japanese speakers \cite{yasuta2004stop}. Our study implies that the production of aspirated stops may also be related to this phenomenon.

Error patterns can also be attributed to differences in \textbf{syllable structure}, especially in the syllable-final position of languages \cite{he2014production}. While Korean allows \textipa{/l/} as coda, four languages excluding English do not allow syllable-final \textipa{/l/}. Although English allows \textipa{/l/} as the final consonant, the allophone \textipa{/\textltilde/} at the word-final position sounds different from Korean \textipa{/l/}. Vietnamese learners especially had a higher chance of mispronouncing \textipa{/l/}, in particular as \textipa{/n/} that has a similar place of articulation \cite{vietnam_l, hwang_viet}. 
Chinese learners show a significant pattern of \textipa{\textipa{[k\textcorner]}} deletion, which was previously explained in \cite{zh_k} with syllable structure. Japanese learners consistently insert \textipa{/\textturnm/} as well, due to their syllable-based constraints with the tendency to open-syllabification \cite{jp_insertion}.

Lastly, \textbf{pronunciation rule} can explain the frequent deletion of coda \textipa{[t\textcorner]}. The corresponding rule is tensification, where the plain consonants are pronounced as tense in a stop-plain consonant sequence. While the sequence must be acoustically realized as long closure, many L2 Korean learners are reported to produce shorter closures compared to the natives, because they are either unaware of or unfamiliar with the tensification rule \cite{lee2018Korean, Kim2019coda}. The mispronounced sounds are often perceived as deletion of coda \cite{lee2018Korean}, which is observed in our automatic phone transcriptions as well. 

%% file: sections/5_conclusion.tex
\input{figures/comparative.tex}
\section{Conclusion} \label{sec:conclusion}
This paper presents a large-scale linguistic analysis on L2 Korean pronunciation error patterns from speakers of five different L1s, Chinese, Vietnamese, Japanese, Thai, and English, by using the automatic phonetic transcription. Comparative analyses regarding L1 background reveal common error patterns and L1-dependent error patterns, which can be further explained by three types of L1 influences: phoneme inventory, syllable structure, and pronunciation rule.
This study contributes not only in integrating and expanding the previous knowledge of L2 Korean error patterns on a large-scale basis, but also validating the possibilities of automatic phonetic transcriptions on non-native speech analysis.
Future work includes considering L1 and uncategorizable phones to yield a better and deeper understanding of L1-influenced non-native pronunciation error patterns.

%% file: figures/comparative.tex
\begin{table}[t!]
\caption{
Frequency values (\%) of confusion matrices. * < .05, ** < .01, *** < .001 
}
\label{tab:compare}
\centering
\resizebox{0.4\textwidth}{!}{
\begin{tabular}{c|c|c|c|c|c}
\hline
\textbf{L1} & \textbf{Canon} & \textbf{Real} & \textbf{L1 Freq. (\%)} & \textbf{Average Freq. (\%)} & \textbf{P-value} \\ \hline
ZH & \textipa{\textipa{[k\textcorner]}} & \textipa{\textipa{[k\textcorner]}} & 58.87 & \textbf{74.95} & * \\ \cline{2-6}
& \textipa{\textipa{[k\textcorner]}} & * & \textbf{22.79} & 7.32 & ** \\ \cline{2-6}
 &  \textipa{/ju/}   & \textipa{/jo/}  & \textbf{23.50} & 11.61 & * \\
\hline
VI & \textipa{/\subdoublevert{\texttctclig}/} & \textipa{/\texttctclig\textsuperscript{h}/} & \textbf{18.73} & 7.99 & * \\ \cline{2-6}
& \textipa{/l/} & \textipa{/l/} & 56.54 & \textbf{73.23} & * \\ \cline{2-6}
& \textipa{/l/} & \textipa{/n/} & \textbf{23.20} & 7.52 & ** \\ \hline
JP & \textipa{/p\textsuperscript{h}/} & \textipa{/p\textsuperscript{h}/} & 51.22 & \textbf{70.70} & ** \\ \cline{2-6}
& \textipa{/p\textsuperscript{h}/} & \textipa{/\subdoublevert{p}/} & \textbf{20.72} & 7.33 & * \\ \cline{2-6}
& \textipa{/t\textsuperscript{h}/} & \textipa{/t\textsuperscript{h}/} & 44.40 & \textbf{62.24} & * \\ \cline{2-6}
& \textipa{/t\textsuperscript{h}/} & \textipa{/\subdoublevert{t}/} & \textbf{20.51} & 5.99 & ** \\ \cline{2-6}
& \textipa{/k\textsuperscript{h}/} & \textipa{/k\textsuperscript{h}/} & 54.21 & \textbf{70.20} & * \\ \cline{2-6}
& \textipa{/k\textsuperscript{h}/} & \textipa{/\subdoublevert{k}/} & \textbf{31.15} & 12.74 & ** \\ \cline{2-6}
& \textipa{/\ng/} & \textipa{/\ng/} & 55.22 & \textbf{82.67} & *** \\ \cline{2-6}
& \textipa{/\ng/} & \textipa{/n/} & \textbf{31.22} & 10.10 & *** \\ \cline{2-6}
& * & [\textfishhookr] & \textbf{10.43} & 1.17 & * \\ \cline{2-6}
& * & \textipa{/\textturnm/} & \textbf{8.98}  & 0.07 & **  \\ 
\cline{2-6}
& \textipa{/w\textturna/} & \textipa{/\textturna/}  & \textbf{14.58}  & 3.96 & *  \\ \cline{2-6}
& \textipa{/w\textepsilon/} & \textipa{/\textepsilon/} & \textbf{26.69} & 9.42 & **  \\ \cline{2-6}
& \textipa{/wi/} & \textipa{/wi/} & 43.28 & \textbf{58.58} & *  \\ \cline{2-6}
& \textipa{/wi/} & \textipa{/i/} & \textbf{32.05} & 11.49 & *** \\ \cline{2-6}
& \textipa{/u/} & \textipa{/u/} & 64.03 & \textbf{77.79} & * \\ \cline{2-6}
&  \textipa{/u/} & \textipa{/\textturnm/} & \textbf{24.72} & 7.08 & ** \\ \cline{2-6}
& \textipa{/\textturnv/} & \textipa{/\textturnv/} & 66.99 & \textbf{81.78} & * \\ \cline{2-6}
& \textipa{/\textturnv/} & \textipa{/o/} & \textbf{25.58} & 10.46 & **  \\ 
\hline
TA & \textipa{/\texttctclig\textsuperscript{h}/} & \textipa{/s/} & \textbf{22.05} & 7.33 & ** \\ \cline{2-6}
\hline
\end{tabular}
}
\vspace{5mm}

\end{table}